\documentclass[10pt,journal,compsoc]{IEEEtran}

\usepackage{booktabs}

\ifCLASSOPTIONcompsoc
  \usepackage[nocompress]{cite}
\else
  \usepackage{cite}
\fi

\usepackage{subcaption}
\usepackage{multirow}
\usepackage{xcolor}
\usepackage{soul}
\usepackage{url}
\usepackage{makecell}
\usepackage{array}
\newcolumntype{L}[1]{>{\raggedright\let\newline\\\arraybackslash\hspace{0pt}}m{#1}}
\newcolumntype{C}[1]{>{\centering\let\newline\\\arraybackslash\hspace{0pt}}m{#1}}
\newcolumntype{R}[1]{>{\raggedleft\let\newline\\\arraybackslash\hspace{0pt}}m{#1}}

\ifCLASSINFOpdf
  \usepackage[pdftex]{graphicx}
\else
\fi

\usepackage{amsmath}

\begin{document}
%
\title{REWIND Dataset: Privacy-preserving Speaking Status Segmentation from Multimodal Body Movement Signals in the Wild}
%
%
%
%

\author{Jose~Vargas-Quiros,~\IEEEmembership{Non-Member,~IEEE,}
        Chirag~Raman,~\IEEEmembership{Non-Member,~IEEE,}
        Stephanie~Tan,~\IEEEmembership{Non-Member,~IEEE,}
        Ekin~Gedik,~\IEEEmembership{Non-Member,~IEEE}
        Laura~Cabrera-Quiros,~\IEEEmembership{Non-Member,~IEEE}
        and~Hayley~Hung,~\IEEEmembership{Member,~IEEE,}
\IEEEcompsocitemizethanks{\IEEEcompsocthanksitem J. Vargas, C. Raman and H. Hung are with the Department of Intelligent Systems at TU Delft, The Netherlands.\protect\\
E-mail: {j.d.vargasquiros, c.a.raman, h.hung}@tudelft.nl
\IEEEcompsocthanksitem L. Cabrera Quiros works at Escuela de Ingenieria Electronica at the Instituto Tecnologico de Costa Rica, Costa Rica.\protect\\
E-mail: lcabrera@itcr.ac.cr}

\thanks{Manuscript received April 19, 2005; revised August 26, 2015.}}

%
%


\IEEEtitleabstractindextext{%
\begin{abstract}
Recognizing speaking in humans is a central task towards understanding social interactions. Ideally, speaking would be detected from individual voice recordings, as done previously for meeting scenarios \cite{Carletta2006}. However, individual voice recordings are hard to obtain in the wild, especially in crowded mingling scenarios due to cost, logistics, and privacy concerns \cite{Raman2022b}. As an alternative, machine learning models trained on video and wearable sensor data make it possible to recognize speech by detecting its related gestures in an unobtrusive, privacy-preserving way. These models themselves should ideally be trained using labels obtained from the speech signal. However, existing mingling datasets do not contain high quality audio recordings. Instead, speaking status annotations have often been inferred by human annotators from video, without validation of this approach against audio-based ground truth. In this paper we revisit no-audio speaking status estimation by presenting the first publicly available multimodal dataset with high-quality individual speech recordings of 33 subjects in a professional networking event. 
We present three baselines for no-audio speaking status segmentation: a) from video, b) from body acceleration (chest-worn accelerometer), c) from body pose tracks. In all cases we predict a 20Hz binary speaking status signal extracted from the audio, a time resolution not available in previous datasets. In addition to providing the signals and ground truth necessary to evaluate a wide range of speaking status detection methods, the availability of audio in REWIND makes it suitable for cross-modality studies not feasible with previous mingling datasets. Finally, our flexible data consent setup creates new challenges for multimodal systems under missing modalities.
\end{abstract}

}

\maketitle

\IEEEdisplaynontitleabstractindextext
\IEEEpeerreviewmaketitle

\IEEEraisesectionheading{\section{Introduction}\label{sec:introduction}}

\begin{figure*}[!ht]
     \centering
     \begin{subfigure}[t]{0.44\textwidth}
     \vskip 0pt
         \centering
         \includegraphics[width=\textwidth]{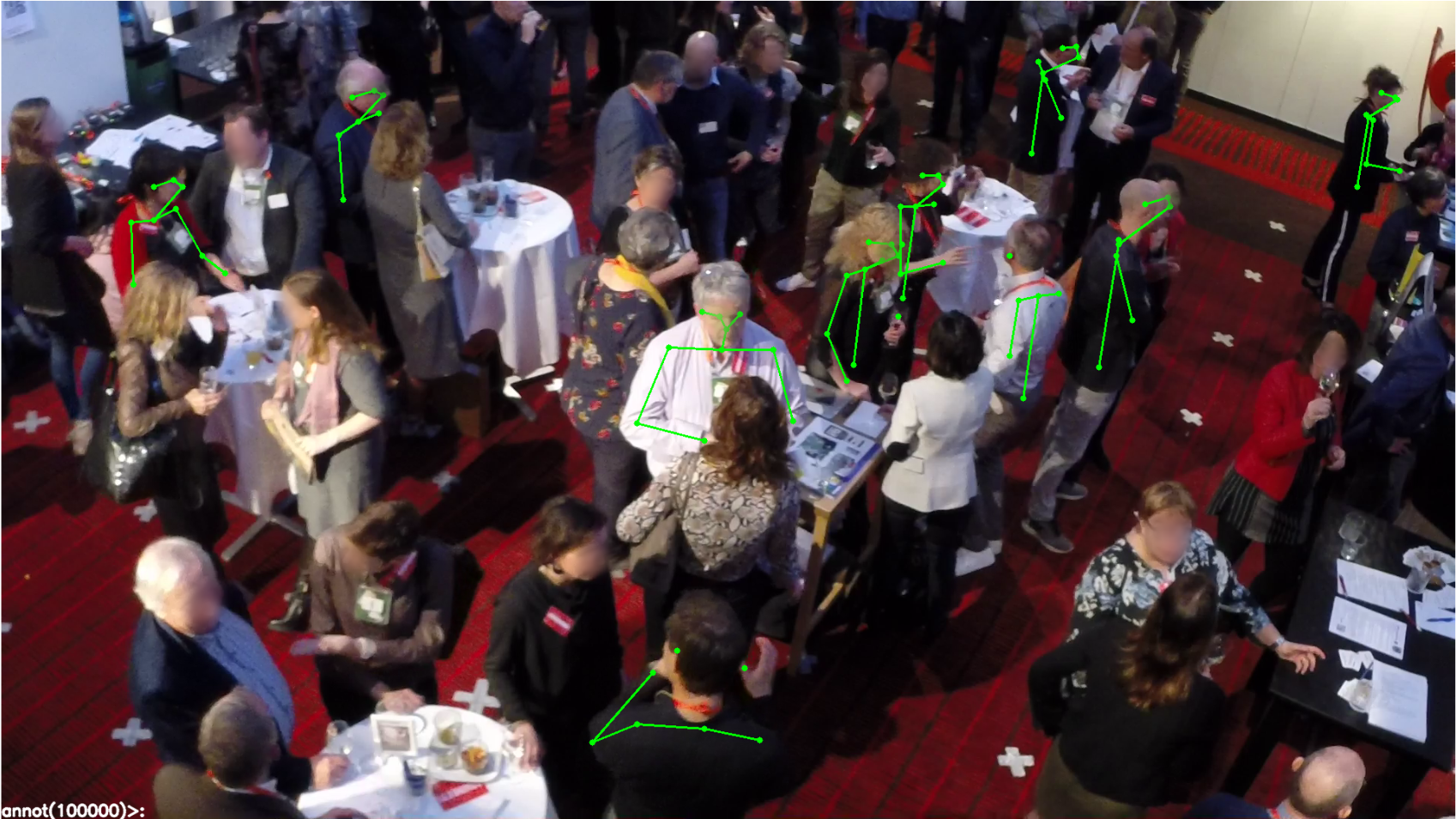}
         \caption{Capture from one of our cameras, including the pose tracks active in the current frame. Note that we only extracted poses for subjects who wore microphones.}
         \label{fig:y equals x}
     \end{subfigure}
     \hfill
     \begin{subfigure}[t]{0.295\textwidth}
     \vskip 0pt
         \centering
         \includegraphics[width=\textwidth]{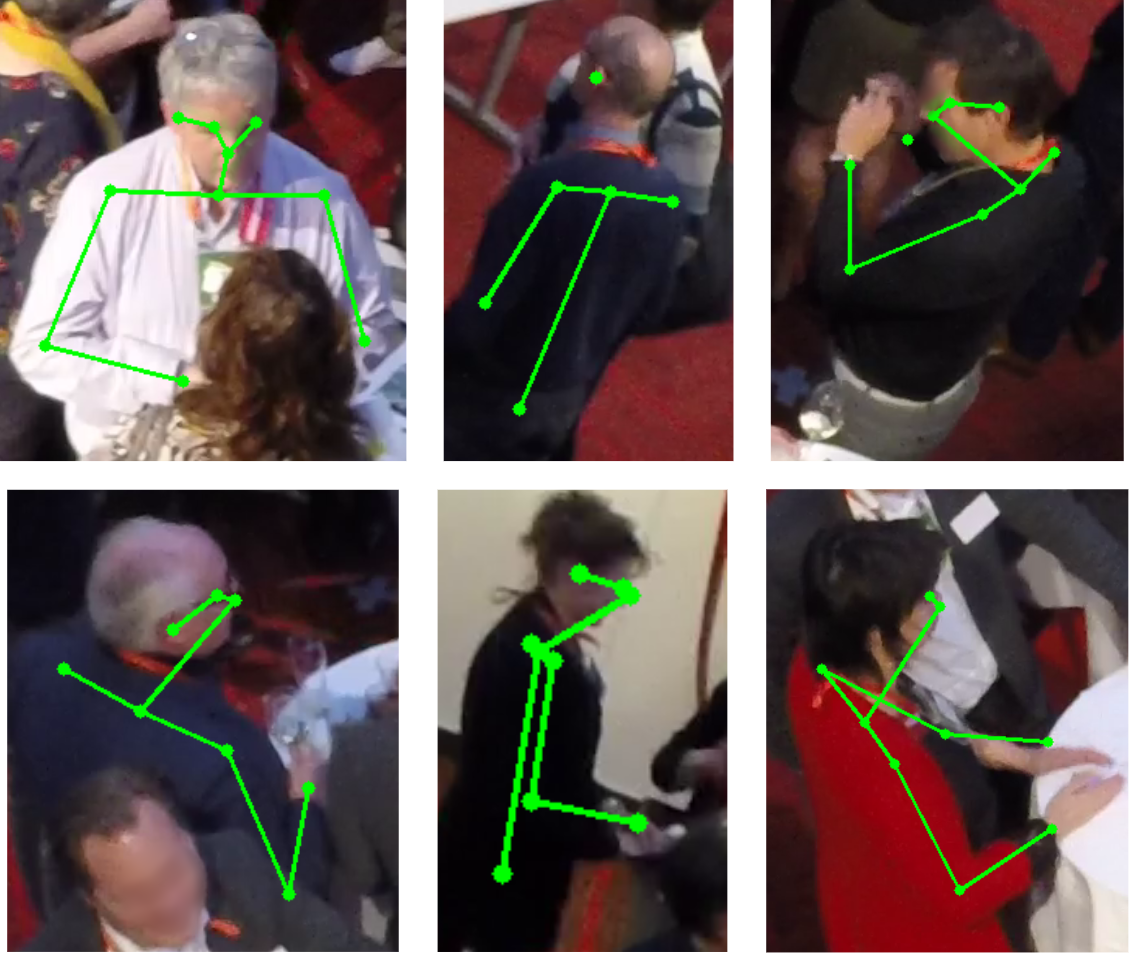}
         \caption{Crops showing individual subject pose tracks.}
         \label{fig:three sin x}
     \end{subfigure}
     \hfill
     \begin{subfigure}[t]{0.168\textwidth}
     \vskip 0pt
         \centering
         \includegraphics[width=\textwidth]{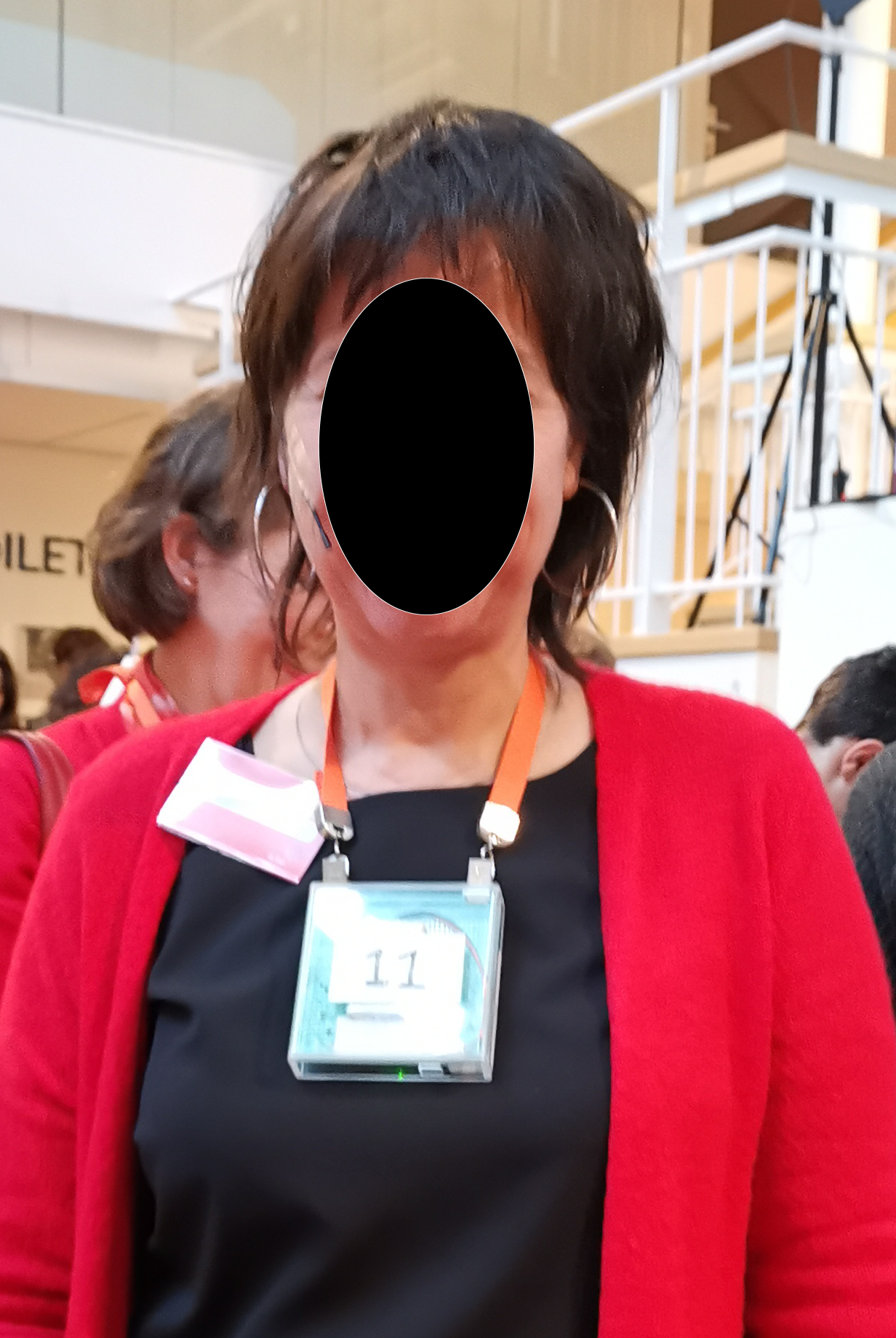}
         \caption{Subject wearing the accelerometer and microphone.}
         \label{fig:dataset:subject}
     \end{subfigure}
        \caption{Captures of our dataset and data subjects.}
        \label{fig:dataset}
\end{figure*}

\IEEEPARstart{D}{etection} or segmentation of speaking activity in free-standing social settings is a core necessity in building systems capable of interpreting interactions in everyday situations, from networking events to exchanges around the coffee machine at the office. The analysis of a complex conversational scene where dozens of people stand, walk, form groups and converse freely (see Fig. \ref{fig:dataset}) is of particular interest in fields such as computational social science and social signal processing for the development of socially intelligent systems capable of aid  \cite{Hung2019a}. Segmenting speaking status (ie. binary signal indicating voice activity of a target speaker) with time resolutions that are suitable for indicating back-channels is key because of its utility in downstream tasks where it can be used, for example, in the quantification of individual and group measures of experience in conversation like involvement \cite{Oertel2011}, satisfaction \cite{Lai2018}, perceived quality \cite{Raman2022, Prabhu2020a}, or affect \cite{Lai2013}, and in the forecasting of future events like speaking, gesturing and changes in position and orientation \cite{Joo2019,Raman2022a}.


The audio modality is the obvious choice for the measurement of speaking status. High-quality speaking status signals have been obtained from personal head-mounted and directional microphones in seated meetings \cite{Carletta2006}. However, individual audio is especially hard to acquire in a mingling setting, or crowded conversational scene. Microphone equipment is hard to scale to large dynamic crowded scenes with synchronization guarantees. Furthermore, recording audio is more likely to raise privacy concerns with event attendees or event organizers. It is our experience from prior data collection efforts that the perception of audio recording, even when using more privacy-preserving low frequencies is a deterrent for participant recruitment. In fact, none of the datasets created for the study of free-standing conversations contain raw high frequency audio \cite{Cabrera-Quiros2018a, Raman2022b}.

Instead, wearable sensors like the sociometric badge \cite{Choudhury2003, Alameda-Pineda2016} and mobile phone sensing \cite{Martella2015} have been used in mingling settings to capture lower-fidelity signals that obscure the content of conversations but may still capture speaking status \cite{Lederman2017}. However, the noise introduced by the large number of sound sources in a crowd makes it challenging to distinguish speakers in a low frequency recording \cite{Alameda-Pineda2016}.

For these reasons, the possibility of detecting speaking from body movement alone without access to audio offers an appealing privacy-sensitive solution to these problems. It has long been observed that hand and head gestures frequently co-occur with speech \cite{Lascarides2009, McNeill1994} while being salient cues with similar motion characteristics across people.

The video modality commonly present in most in-the-wild mingling datasets \cite{Cabrera-Quiros2018a, Alameda-Pineda2016, Raman2022b} offers a convenient way to observe these gestures, with several methods having been proposed to detect speaking status from video \cite{Cabrera-Quiros2020, Vargas2019a, Wanga}. So far, however, none of these methods have been trained using ground truth obtained from high-quality individual audio recordings. Video-based annotation or noisy low-frequency signals have been used instead. While video-based annotations of speaking status do provide a valid supervisory signal, it is currently unclear if audio-based labels would provide higher quality annotations given that shorter turns such as back-channels are not so easily observable. Having access to reliable back-channel annotations is an important step for social involvement detection, for example. 

In this work, we present the first dataset for the study of speaking status segmentation from body movement in the wild with raw, high-quality audio from personal directional Lavalier-type microphones. REWIND offers for the first time, the possibility to scrutinise the relationship between speech and body movements in this setting, with audio-based ground truth that is more fine-grained (higher temporal resolution) compared to previous datasets. Beyond speaking status, the more general problem of how body movement can be used to infer phenomena observed/annotated in the audio modality opens the door for the cross-modal study of other social cues and signals like laughter and back-channeling. In fact, the REWIND dataset was already used in \cite{Vargas-Quiros2022} to study laughter annotation across modalities.

In addition to audio, the dataset includes three modalities capturing body movements: video, pose, and wearable acceleration. Video recordings include top-down and side-elevated views. The latter was used to automatically obtain pose tracks for subjects in the scene. Acceleration readings were obtained from wearable devices in a badge-like form factor worn by data subjects on the chest.


Our contributions are the following:

\begin{itemize}
    \item We introduce the REWIND dataset, the first in-the-wild mingling dataset with high-quality raw audio, video, and acceleration; automatic pose annotations, and automatic speaking status labels. 
    
    \item We present results from four body-movement-based speaking status segmentation (SSS) machine learning tasks: 1) video-based SSS, 2) acceleration-based SSS and 3) pose-based SSS, and 4) multimodal (video + acceleration + pose) SSS. In comparison with previous work, we increase the resolution of our outputs by training our models to produce a probability mask for speaking status over an input segment. 
    
    \item We analyze the role of the REWIND dataset in the context of speaking status segmentation from body movement, and in the wider field of the computational study of body movements, identifying potential research directions enabled by a dataset such as REWIND.
\end{itemize}

\section{Related Work}\label{sec:related}

\begin{table*}[t!]
\caption{Mingling datasets with speaking status labels or related data.}
\label{tab:dataset-comparison}

\scriptsize \setlength{\tabcolsep}{3pt}
\begin{tabular*}{\textwidth}{@{\extracolsep{4pt}}lC{1cm}L{1cm}L{1.6cm}L{1.6cm}L{3cm}L{3cm}l@{}}
 & \multicolumn{2}{c}{} & \multicolumn{2}{c}{Audio and Speech Labels} & \multicolumn{3}{c}{Body Movement Modalities}\\
 \addlinespace[0.05cm]\\
\cline{4-5}\cline{6-8}\\
Dataset  & \makecell[c]{\# Subj} & Length & Audio & VAD labels & Video & Pose tracks & Acceleration \\
\addlinespace[0.05cm]\\
\cline{1-8}\\
MatchNMingle \cite{Cabrera-Quiros2018a} & 92 & 30 min & No & From video & Top down & No, only BBs & Yes \\
\addlinespace[0.2cm]
SALSA \cite{Alameda-Pineda2016} & 18 & - & STFT @ 30Hz & From video & Side-elevated & No, only BBs & Yes \\
\addlinespace[0.2cm]
ConfLab \cite{Raman2022b} & 48 & 16 min & 1250Hz & From video & Top-down & Yes & Yes (IMU) \\
\addlinespace[0.2cm]
\textbf{REWIND (ours)} & 18 & 90 min & 44100Hz & From audio & Top-down \& side-elevated & Yes & Yes \\

\addlinespace[0.2cm]

\cline{1-8}\\
\end{tabular*}
\end{table*}

Although generic action recognition and localization tasks have received the most attention in the literature \cite{Kong2022}, some work has been concerned specifically with speaking status detection or segmentation without access to audio \cite{Beyan2020, Beyan}, and the challenges specific to in-the-wild mingling settings \cite{Wanga, Cabrera-Quiros2018, Vargas2019a, Cristani2012, Gedik2017, Chen2006, Matic2013}. Note that the terms \textit{detection} and \textit{segmentation} may be used arbitrarily for models operating across a wide range of time resolutions, and we use them interchangeably while underscoring the importance of higher resolutions in supporting a wider variety of research questions. In this section we review detection and segmentation work, starting with the datasets and methods most related to our scenario. Additionally, we discuss speaking status methods developed for settings other than in-the-wild mingling, and how they fail to address key challenges specific to in-the-wild mingling.

\subsection{Related Datasets and Methods in Mingling Settings}
\label{sec:related_mingling}

In the study of conversational social behavior, turn-taking patterns are a fundamental unit of analysis. Therefore, most previous mingling datasets have contained speaking status annotations and have presented estimating it as a baseline task \cite{Cabrera-Quiros2018a, Alameda-Pineda2016, Raman2022b}. These datasets contain raw modalities like video and acceleration from wearable devices. Subjects are usually localized in videos using bounding boxes, with the exception of the recent ConfLab dataset \cite{Raman2022b} which contains full-body keypoint annotations. Table \ref{tab:dataset-comparison} presents an overview of existing mingling datasets used for speaking status detection. Follow-up work using some these datasets has addressed speaking status detection from video by classifying 3-second windows as speech / non-speech \cite{Cabrera-Quiros2020, Vargas2019a, Wanga, Gedik, Vargas-Quiros2022a}. Cabrera-Quiros et al. \cite{Cabrera-Quiros2020} presented MILES, a method making use of multiple instance learning to classify bags of dense trajectories, also making use of 3-second windows. In a MediaEval multimedia evaluation benchmark addressing this task, Fisher Vectors were also explored as an alternative to represent dense trajectories \cite{Vargas2019a}. Wang et al. showed significant improvements over both of these methods using a 3D CNN method on 1-second windows.

Poses, often derived from video frames, are another modality of interest due to their lower dimensionality compared to raw video and their ability to capture gestures. Individual pose detections have been used in action recognition work, both as standalone inputs and in combination with video, to provide precise localization information \cite{Choutas2018, Yan2019}. However, the application of this work to speaking status detection in the wild has been limited \cite{Vargas-Quiros2022a}. It is not known whether body-pose-based methods are capable of reaching the same levels of performance as video-based ones, or the effect that different pose detection approaches have on performance.

Despite their advantages, video and pose inputs to action recognition models are affected by subject occlusion, cross-contamination, poor lighting conditions, and differences in perspective, orientation, and distance with respect to the camera. Wearable accelerometers circumvent the aforementioned challenges and are capable of capturing subtle body movements in space. 
Accelerometer readings from a smart ID badge hung around the neck (Figure \ref{fig:dataset:subject}) have long been studied for recognizing actions in mingling settings \cite{Hung2014, Hung2013a}. Hung et al. \cite{Hung2013a} explored recognition of actions like gesturing, laughing, and speaking, obtaining the highest performance among them for speaking detection. The MediaEval multimedia evaluation benchmark also evaluated acceleration-based methods \cite{Gedik} on a subset of the MatchNMingle dataset \cite{Cabrera-Quiros2018a}, which used the same accelerometer sensors used in this work. Here, CNNs for time series have been shown to improve over traditional classifiers \cite{Vargas2019a}. Particularly, a previous approach made use of transfer-learning to obtain person-specific classifiers \cite{Gedik2017}. Multiple works have found speaking status detection performance from acceleration to be higher than that of video-based methods. \cite{Gedik2017, Cabrera-Quiros2020, Vargas2019a, Raman2022b} As with video, all of these works made use of 1-second or 3-second windows for classification.


Despite all these works, a key challenge remains: existing mingling datasets \cite{Cabrera-Quiros2018a, Alameda-Pineda2016, Raman2022b} do not contain high-quality audio. In \cite{Cabrera-Quiros2018a} and \cite{Raman2022b} speaking status has been annotated from the video. In \cite{Alameda-Pineda2016} speaking status was annotated from low-resolution audio and the authors noted the difficulty of distinguishing speakers in their recordings. Lack of high-quality audio in these datasets not only limits the time-resolution of the speaking status annotations (and of the trained models) but may affect the correctness of the annotations themselves. The quality of video-based speaking status annotations has not been compared with that of audio-based ground truth. Furthermore, lack of audio makes such datasets unusable for the study of other verbal phenomena, due to the impossibility of annotating them.

\subsection{Speaking Status Detection in Non-mingling Settings}

In work not specific to in-the-wild mingling settings, researchers have addressed the problem of speaking status detection / segmentation (also termed Voice Activity Detection), especially from upper-body shots of people in videos \cite{Shahid2019}. Beyan et al. \cite{Beyan2020} introduced the RealVAD dataset, which consists of a single-camera frontal recording of a panel discussion where 9 subjects take turns speaking. The RealVAD method presented in the same paper adapts CNN-extracted features from one speaker to another to improve performance. Previous work presented the Columbia dataset \cite{Chakravarty2016a}, set in a similar panel discussion setting. These datasets contain higher resolution, audio-based labels compared to mingling-specific datasets (Section \ref{sec:related_mingling}). However, they have some fundamental differences with the mingling setting. In particular, freedom of movement in mingling datasets creates the challenge of learning from data points with a variety of camera angles, occlusion levels, orientations, and distances (of subjects respective to the camera) not present in panel discussion datasets. Perhaps due to the absence of occlusion challenges, the use of wearable acceleration has not been explored in these datasets. Furthermore, panel discussions have specific dynamics with most often a single speaker at a time.  Another related task is speaker naming in movies \cite{Hu2015, Roth2020}. However, here it is normally assumed that the algorithm has access to audio.


\section{Data Acquisition}\label{sec:acq}

Investigating speaking status in a naturalistic setting involves the collection of a dataset in which social interaction occurs with as little intervention as possible. Following the design principles outlined in \cite{Raman2022b}, we collected the dataset in collaboration with organizers of a special event for a business networking group. In this section, we detail the data collection procedure and setting in which our data was collected (\ref{sec:acq_setting}) along with our sensor setup (\ref{sec:acq_sensor}).

\subsection{Participant Procedure}\label{sec:acq_setting}

Most participants in the networking group met regularly and many but not all of them knew each other. Participants were informed beforehand that this particular meeting would be recorded. As they arrived at the event, attendees were approached one by one and informed again of the special circumstances of the data collection. They were then informed of the data collection process and invited to donate their data. They were free to choose which sensors to wear between microphone, accelerometer, or both; or to not participate at all. They were then asked to sign an Informed Consent Form. Subsequently, participants were fitted with the corresponding sensors. To enable the possibility of opting out of the video modality, all participants were informed about a clearly-delimited video zone where they would be recorded by our video cameras. 

After this, subjects were free to move around the room and talk as they pleased. During the first half of the event (1.5hr), however, they were at times expected to attend to a speaker, a live music performance, and participate in social games and activities. In the second half of the event (1.5hr) subjects were free to mingle without interruption, as there were no more games or activities. The room was not closed and they were also free to leave at any time, after returning their sensors. During both halves of the event, most of the interaction consisted of free-standing conversation, as there was little seating available. The mood appeared friendly and relaxed.

When participants approached to return their sensors, we asked them to fill an exit survey indicating their experience in the event, including rating on a scale of 1-5 their perceived level of enjoyment ($4.14 \pm 0.79$), their likelihood of attending an event like that one again ($4.21 \pm 0.70$) and of recommending the event to others ($4.12 \pm 0.72$), and free-form textual feedback. The survey was associated to their sensor IDs. After the event, we sent the subjects a report of their behavior, which included information (relative to the other subjects) about their speaking time, amount of motion and number of interaction partners.

Our complete data collection process, as outlined above, was approved by the ethics board of Delft University of Technology beforehand. 

\subsection{Sensor setup}\label{sec:acq_sensor}

We collected the following data from consenting participants:

\begin{LaTeXdescription}
	\item[Audio] Lavalier microphones attached to the face using lavalier tape\footnote{ Lavalier tape, or LAV tape is an extremely fine tape that is designed to cause minimal restriction to facial muscle movements. It is transparent and is used by professional theatre productions so that the microphone remains as inconspicuous as possible. This was an important design consideration to minimise discomfort and visual distractions on the participant's face}, recorded speech at 44KHz. Microphones were attached to a Sennheiser SK2000 transmitter attached near the subject's waist. Transmitters communicated wirelessly with a central receiver, which stored fully synchronized audio in real time.
	\item[Body acceleration] A custom-made wearable tri-axial accelerometer sensor was hung around the neck and rested on the chest like a smart ID badge (Figure \ref{fig:dataset:subject}), recording at 20Hz.
	\item[Video] 12 overhead cameras and four side-elevated cameras were placed above and in the corners of a video zone. Every camera recorded video at a resolution of $1920x1080$ and 30fps. In this work we only make use of the four side elevated cameras.
\end{LaTeXdescription}

Data was collected from the moment the participant received the sensors to the moment they returned them, which varied from about 30 minutes to more than 3 hours for some subjects who stayed after the event was officially over.

\subsection{Data Collection Details}\label{sec:dataset_stats}

Because some participants chose to only wear one of the sensors or to not enter the video zone, and because of the malfunction of some of our wearable devices, not all modalities were available for all participants. Of about 100 attendees to the event, 33 wore a microphone and 52 wore an accelerometer; while 25 wore both sensors. Most of the participants interacted within the video zone. 

Our recordings included segments when the participants were expected to listen to a speaker or a performance (Section \ref{sec:acq_setting}). We used the videos to manually find these segments and exclude them from our experiments as they deviate from our setting of interest.

\section{Data Annotation}\label{sec:annot}

In this section, we explain how the speaking status labels were generated using the individual audio recordings (Section \ref{sec:annot_vad}) and semi-automatically for poses from side-elevated videos (\ref{sec:annot_pose}). 

\subsection{Automatic Audio-based Speaking Status Annotation}\label{sec:annot_vad}

Speaking status is generally labeled as a binary variable, where a positive value indicates voice activity from a target subject. The availability of high-quality audio recordings from head-worn microphones allowed us to automatically obtain labels for speaking activity via speech processing algorithms. Our task is particular in the amount of background (cocktail party) noise present and interlocutor speech present in the recordings, greater than in many speech datasets which use meeting or phone recordings \cite{Carletta2006, Carletta2006}. This includes interlocutor speech from subjects close to the microphone wearer, whose words could even be understood clearly from the recordings. Our goal was to ignore both interlocutor speech and background noise, and obtain labels indicating speaking status of the microphone wearer only.

We initially evaluated two VAD approaches: the \textit{pyannote.audio} package \cite{Bredin2020, Lavechin2020} and rVAD method \cite{Tan2020}). However, we found to be unsuccessful in dealing with the noise in our audio recordings. We therefore decided to use a denoising step followed by a diarization step via \textit{pyannote.audio} and \textit{NVIDIA NeMo} libraries respectively. The complete process to generate VAD labels is a follows:

\begin{enumerate}
    \item Loudness normalization (EBU R128) to normalize differences in audio energy across recordings (which could be due to microphone fit). We used ffmpeg's \textit{loudnorm} filter \cite{FFmpegDevelopers2016}. 
    \item Denoising via \textit{Speechbrain}'s SepFormer model \cite{subakan2021attention} trained on the WHAM! dataset \cite{Wichern2019WHAM}. We ran the method using a 1-minute sliding window due to model input size limitations. This removed most of the cocktail party noise in the data, but not the voices of interlocutors.
    \item Speaker diarization via NVIDIA NeMo \cite{kuchaiev2019nemo}, a pipeline including VAD, segmentation, speaker embeddings, and clustering. We found the method to effectively separate the wearer's voice in its own cluster, distinguishing it from other speakers.
    \item We manually identify the microphone wearer's cluster in the diarization outputs, with help from video recordings to identify the speaker. We transform the speech segments into a speaking status time series using their timing information.
\end{enumerate}

\subsection{Semi-automatic Pose Annotations}\label{sec:annot_pose}

Since most pose estimation algorithms, including OpenPose, work independently on individual frames, we needed an approach to associate poses across frames. This problem has been investigated in previous work such as PoseFlow \cite{Xiu2019}. However, we found this method to be too computationally expensive for our use case due to the large number of people in the scene. We therefore implemented a semi-automatic method to obtain tracks from individual frame detections. We chose a computationally lighter method based on the observation that the chest keypoint was reliably detected and localized across frames. 

Specifically, our goal is to create pose tracks by associating skeletons or poses across frames. We chose to do so by iterating over frames and assigning the poses detected in each frame to existing or new tracks. Specifically, for each frame $n$ of the video, the pose detector outputs a set of poses given by $Q_n = \{P_{n,m} \mid m = 1, \dots, M_n\}$, where $M_n$ is the number of people detected in frame $n$ and $P_{n,m} = \{\pmb{p}_{n,m,j} \mid j = 1, \dots, J\}\text{; }\pmb{p}_{n,m,j} = (p_{n,m,j}^{x}, p_{n,m,j}^{y})$ is a vector of $J$ 2D-keypoints (or joints) representing a skeleton in the image plane. A pose track is a sequence of poses given by $U_{i,f} = \{P_{i,m}, \dotsc, P_{f,m}\}$, starting at frame $i$ and ending at frame $f$.
We associate poses in $Q_n$ to tracks in $T_n$, the set of open tracks (consisting of poses from all frames up to $n-1$). We do so by comparing poses in $Q_n$ with the head of existing tracks whose last detected pose is not older than $R_{th}$ frames, where $R_{th}$ is an integer threshold parameter. In other words, we solve the assignment problem between two sets of poses: $Q_n$ and $\{P_{f, m} | \forall U_{i,f} \in T_n \text{ and } n - f < R_{th}\}$. This process is repeated in order for $n=1,\dotsc,N$.

The assignment problem remains to be solved. We define the distance between two poses as the Euclidean distance between their chest keypoints across frames; ie. for frames $A$ and $B$, $D(P_{n_1, m_1}, P_{n_2, m_2}) = ||\pmb P_{n_1, m_1}^{chest} - \pmb P_{n_2, m_2}^{chest}||$. We solve the assignment problem via the Hungarian algorithm. We add a maximum distance threshold $D_{th}$ for assignment, such that if $D(P_{n_1, m_1}, P_{n_2, m_2}) > D_{th}$, then $P_{n_1, m_1}$ and $P_{n_2, m_2}$ cannot be assigned to each other. Assigned keypoints are added to the corresponding track and unassigned keypoints are assigned to a new track. When a new pose in frame $n_1$ is matched to a pose in a frame $n_2 \neq n_1-1$ (not the immediately preceding frame), we impute the keypoints via linear interpolation to maintain the continuity of the track. 

Parameters $R_{th}$ and $D_{th}$ were set based on a qualitative evaluation of the algorithm on a subset of the tracks. For $R_{th}$, we found a one-second threshold to work best creating consistency across frames without introducing significant errors. This approach resulted in high-quality tracks, with only sporadic track misassignments due to subjects walking in front of one another.

Because our goal was to obtain high quality tracks to be able to reliably test our recognition method (see next sections), we manually inspected the dataset for track switches and corrected them by splitting the tracks. Finally we assigned tracks to subject IDs to be able to associate with the personal acceleration readings.

\section{Dataset Statistics}\label{sec:stats}

\begin{figure}[!ht]
\centering
\includegraphics[width=0.8\columnwidth]{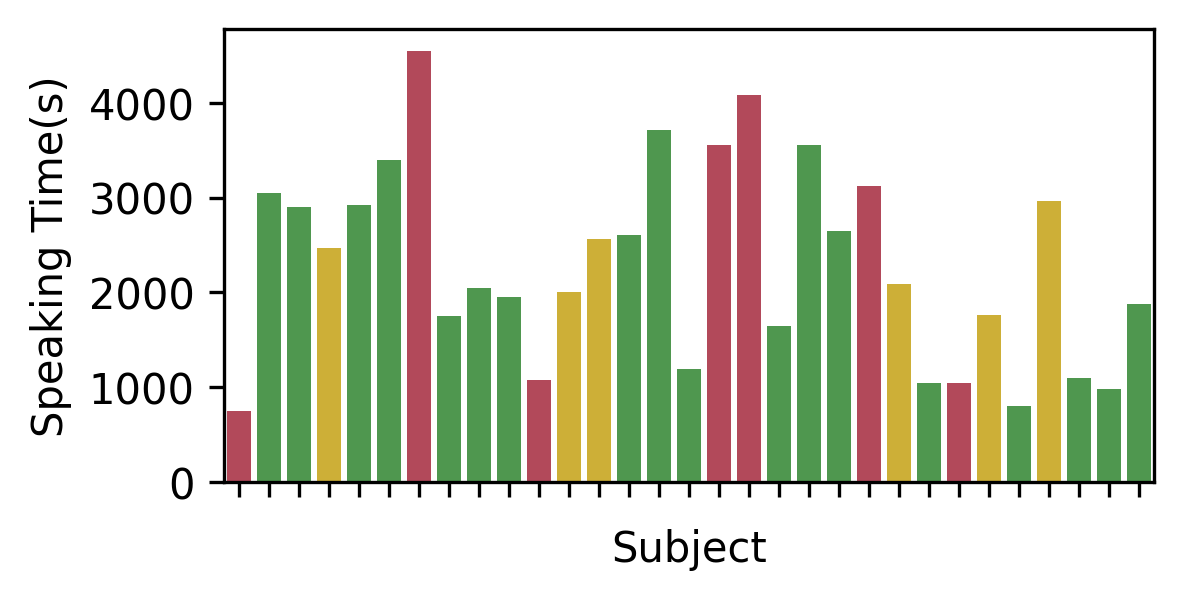}
\caption{Seconds of speaking time per subject with speech data in REWIND dataset. Columns in green indicate subjects with complete information (audio, video, acceleration). Columns in yellow indicate subjects with audio and acceleration information, but who are not visible in the videos (no pose). Columns in red indicate subjects without any body movement information (only audio). }
\label{fig:speaking_times}
\end{figure}

Due to our mixed-consent data collection design, our final dataset contained subjects with different body movement signals available. Subjects were also in the scene for varying amounts of time, and some engaged more than others in conversations. Figure \ref{fig:speaking_times} plots the speaking times per subject, calculated as the summation of all the speaking segments output by VAD (Section \ref{sec:annot_vad}), together with an indication of the modalities available for each subject. The majority of subjects have complete information. There is quite some variation in speaking time with no clear correlation between modality and amount of time spent speaking.

\begin{figure}[!ht]
\centering
\includegraphics[width=0.8\columnwidth]{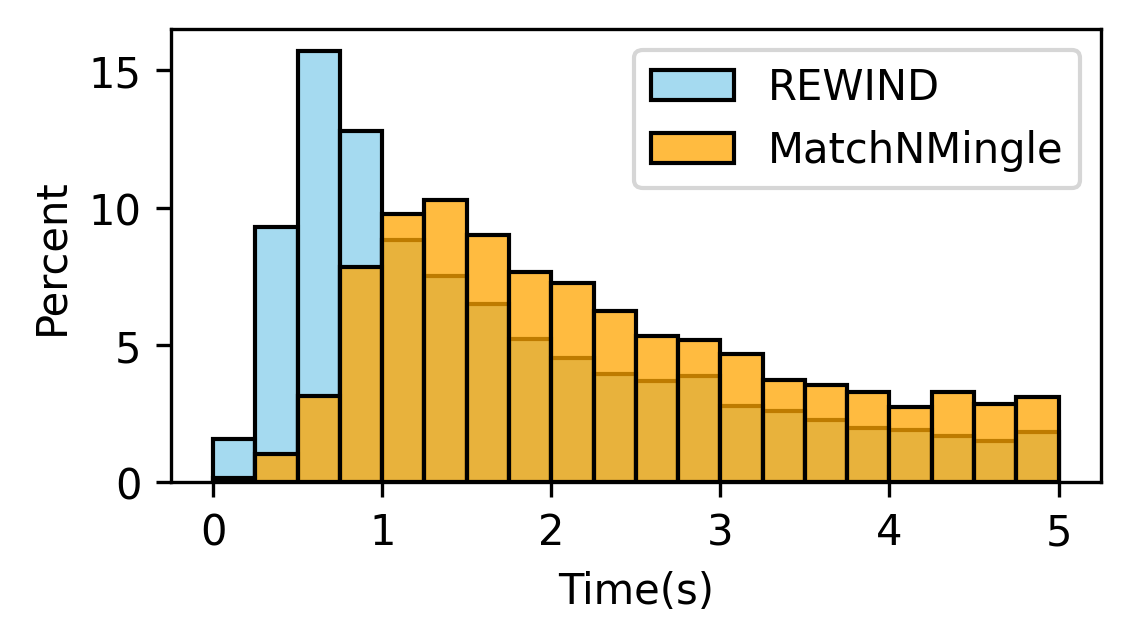}
\caption{Distribution of length of contiguous speaking segments (s) in the speaking status detection labels for REWIND and MatchNMingle \cite{Cabrera-Quiros2020} datasets. REWIND shows greater temporal granularity (shorter segments), thanks to annotations having been obtained from audio. }
\label{fig:turn_lengths}
\end{figure}

To showcase the value of our automatic VAD annotations obtained from audio compared to VAD annotations in previous datasets, we compared the distribution of speaking segments in REWIND to that of the MatchNMingle dataset \cite{Cabrera-Quiros2018}. Figure \ref{fig:turn_lengths} shows the length distribution of segments labeled as speech in both datasets. Although these speaking segments do not constitute turn-lengths, our data contains more short speech segments. Inspection of the dataset revealed these to often correspond to back-channels and short utterances. We interpret that many such utterances were probably missed in previous datasets due to the use of video for speaking status annotation \cite{Cabrera-Quiros2018, Raman2022b}. 

\section{Baselines for Automated Speaking Status Segmentation}\label{sec:baselines}

In this section, we present three baseline tasks for speaking status segmentation from body motion. We start by detailing our evaluation setup, including the process for generating training and testing examples, evaluation metrics, and hyper-parameter tuning. We first present the video baseline, where we train on video patches around target subjects. Second, we present a pose-based approach, where we train on pose track segments. Finally, we present a method based on acceleration readings.

\subsection{Evaluation Setup}

We formulated the problem as speaking status segmentation to take advantage of the high time-resolution of our labels. We defined data samples as 3-second segments of behavior, following previous work which found this length to be close to ideal for speaking status detection \cite{Gedik2017, Cabrera-Quiros2018, Chen2006, Vargas-Quiros2022a}. However, unlike prior works which would predict a single label per 3-second window, our method provides finer granularity by predicting the entire binary speaking status time series (20Hz) for a given data sample.

For consistency across models, we excluded subjects not containing all three modalities (video, pose, acceleration). We leave it to future studies to investigate further the trade-offs. This left us with a total of 18 subjects. We obtained 3-second windows by splitting pose tracks using a sliding window with no overlap. This resulted in a dataset of 16403 examples in total.

For all network hyper-parameters, we used their default values, found to work best over a variety of datasets. For setting the number of training epochs, we used a held-out set of 10\% of the dataset as a validation set. Here, due to the small number of subjects, we partitioned by data point rather than by training subject. Using the rest of the dataset (90\%) we evaluated via 3-fold cross-validation at the subject level, to measure generalization to new subjects. We measured performance via Area under the ROC curve (AUC), where we treat every window element (of which there were 60) as one separate prediction. Although metrics like Intersection over Union (IoU) are often used in segmentation, we made use of AUC due to its robustness against class imbalance and because it is the most conservative estimate for the localisation of speaking samples.

\subsection{Video-based Speaking Status Segmentation}

Due to the relatively small size of our dataset, training state-of-the-art video action recognition methods from scratch would be infeasible. We focused on approaches with pre-trained models available to use as feature extractors. Among those, 3D convolutional neural networks (CNNs) are known to reliably achieve top performances in action recognition benchmarks. We decided to make use of a 3D ResNet pretrained on Kinetics-400, a large action recognition dataset with 400 action classes and over 300000 labeled video clips. The network implementation and models are available as part of the \textit{Pytorchvideo} library \cite{fan2021pytorchvideo}. To adapt the network to our outputs, we implemented a custom network head to apply pooling and convolution operations over the spatial and channel dimensions, and up-sample the time dimension to the length of the target mask (60). Details are in Appendix \ref{app:seg_method_details}.

\subsection{Pose and Acceleration-based Speaking Status Detectors}

For pose and acceleration, we made use of a ResNet variant for time series, implemented as part of the \textit{tsai} library \cite{tsai}. Given the much lower dimensionality of these modalities (when compared to video) we trained both models from scratch. As with the video method, we implemented segmentation network heads to output masks of length 60. Details of the network heads are in Appendix \ref{app:seg_method_details}.

\subsection{Multimodal Speaking Status Segmentation}
Given that our hypothesis about the link between speech and body movement is modality agnostic, it makes sense to also include a baseline combining all three forms of body movement representation together. The assumption here is pose will distinguish head and hand gestures from general body movement, for examples. The video-based will capture more subtle movements that might be obscured by noise from associating the detected pose skeletons between frames as well as shape and texture characteristics of a person's behaviour. Finally, the acceleration will capture sub-pixel and 3D characteristics of the movements that are not discernable in the video. For models using a combination of video, poses, and acceleration inputs, we joined the architectures above by averaging their output masks (output fusion). Network heads were retrained from scratch.

The results of our evaluation are presented in table \ref{tab:results}.
Results suggest the superiority of combining modalities for the task. For video and acceleration, these results align with previous work on speaking status detection which found these modalities to perform comparably \cite{Wanga, Cabrera-Quiros2020}. However, our pose method performed poorly compared to video and acceleration methods. We think that the main reason is likely the noisy nature of the poses. While our approach delivered reasonable track association performance, the fact that tracks are extracted independently per frame introduced significant noise across frames. The relative nature of poses would likely make it harder for the model to separate speech-related gestures from pose noise. It is also possible that making use of pre-training in a state-of-the-art skeleton action recognition method could improve these results. Note, however, that large pre-trained skeleton action recognition methods are often trained on sequences with more than one skeleton, and do not use the same skeleton definition (input size and semantics) \cite{Liu2020}. This makes adapting them to our problem not trivial.


\begin{table}
\centering
\caption{Results of our baseline evaluations.}
\begin{tabular}{|p{3cm}c| } 
\hline
Method & AUC \\ 
\hline
Video 3D-CNN & 0.615 \\ 
Pose CNN & 0.530 \\
Accel. CNN & 0.634 \\
Video + Pose + Accel & 0.648 \\
\hline
\end{tabular}
\label{tab:results}
\end{table}

\section{Discussion and Conclusions}
\label{sec:discussion}

With REWIND, we contribute the first dataset for speaking status segmentation recorded in a real-life mingling scenario and with high-quality individual audio recordings, and derived speaking status annotations. Although the use case of the dataset showcased in this paper is speaking status segmentation, REWIND creates opportunities for research beyond this task, in the automatic detection / segmentation of body movement manifestations of social signals in general. In this section we discuss the implications of the dataset, starting with the results presented in this paper, and following with discussion of the possibilities brought about by REWIND in other related tasks.

\subsection*{REWIND as a Dataset for No-Audio Speaking Status Segmentation.}

Previous speaking status works have shown that it is possible to perform speech/non-speech classification from body movement information \cite{Gedik2017, Cabrera-Quiros2018a, Raman2022b}, as well as classifying the current speaker in a group \cite{Alameda-Pineda2016}. The same lack of audio makes it impossible to verify this, but our results comparing turn length distributions of REWIND with MatchNMingle, suggest that many short ($<1s$) speech segments are either lost or aggregated into longer segments (losing granularity). Furthermore, annotating speech from visual modalities may mean that speech segments were not missed at random, but that the most "visually subtle" speech is missed, resulting in an undesirable bias. A more comprehensive study analagous to \cite{Vargas-Quiros2022} would shed more light on this.

With REWIND, we have presented an approach to record, process, and automatically extract speaking status labels from a small mingling crowd. This approach resulted in a different distribution of speech segment lengths when compared to video-based labels due to a shift towards shorter segments (higher granularity). Furthermore, the time resolution of our labels enables a task not previously attempted: segmentation of speaking status. The availability of ground truth audio means that our annotations are easy to manually verify or further refine automatically in the future. 

One draw-back however, is that the guaranteed level of synchronisation of the REWIND dataset is within latencies of 1s. More precise investigations would likely require estimations over longer window lengths or the collection of new data with high quality audio using a similar multi-sensor synchronisation strategy as \cite{Raman2022b}.

\subsection*{REWIND and the Study of Body Movement.}
REWIND, through its high-quality audio recordings creates opportunities for studying the relationship between vocal production and body movement in naturalistic social interaction. Social actions with a vocal component such as laughter and back-channeling have been studied in the past in relation to body movement \cite{Vargas-Quiros2022, Griffin2013b, Maynard1987}. Higher-level multimodal constructs such as affect, enjoyment or engagement also have manifestations in both vocal production and body movement. With REWIND, in addition to providing aggregate self-reports of enjoyment, we provide the raw data necessary for performing third-party annotations of these constructs from audio, video, or audiovisual information at a higher temporal resolution.

This creates opportunities for using REWIND to train action detectors using different input and labeling modalities. Furthermore, it also allows for the exploration of the effect that different labelling conditions (eg. video-based labeling) have on both label reliability and model performance. This can further our understanding of the trade-offs involved in labelling inherently multimodal phenomena from limited modalities such as video and audio. In fact, a study making use of the REWIND dataset has already addressed such questions in the context of laughter detection, intensity estimation, and segmentation \cite{Vargas-Quiros2022}.

\subsection{Efficacy of Pose-Based Analysis}
One limitation of REWIND lies in the quality of the pose tracks. Due to challenges like occlusion and cross-contamination, pose tracks obtained from our system are noisy, and may miss subjects, especially those far away from the camera. While we consider our tracks to be enough for many applications including evaluation of action recognition methods, they may not be enough for evaluation of tasks like person detection or tracking, where the goal is to detect/track all the people in the frame.

\subsection*{Mixed-Modality Consent: Limitation and Opportunity}
Finaly, the reader may consider that another limitation of REWIND is that many users in the scene did not wear our instruments (Section \ref{sec:stats}). This means that analysis or prediction of social signals from group-level information is difficult with this dataset, with the exception that video data is available for entire groups, and could be used to predict individual variables (eg. speaking status). While this can be seen as a limitation it also affords us an opportunity to investigate such mixed consent settings and the analysis of partially complete data.

\section*{Acknowledgements}
This research is supported by the Netherlands Organization for Scientific Research (NWO) under project number 639.022.606. We acknowledge Bernd Dudzik, Xianhao Ni, Xiang Teng and Alessio Rosatelli for their assistance during the data collection event.

\ifCLASSOPTIONcaptionsoff
  \newpage
\fi



\bibliographystyle{IEEEtran}

\bibliography{library.bib}



\begin{IEEEbiographynophoto}{Jose Vargas} is a PhD candidate at the Socially Perceptive Computing Lab at TU Delft, The Netherlands, since 2018. He is interested in multimodal action recognition and conversation quality assessment in-the-wild, the study of interpersonal adaptation and synchrony, and efficient annotation of in-the-wild data.
\end{IEEEbiographynophoto}

\begin{IEEEbiographynophoto}{Chirag Raman} is a PhD Candidate at the Socially Perceptive Computing and Pattern Recognition Labs at TU Delft, The Netherlands. His research interests include multimodal machine learning, generative modeling, affective computing, computer vision, and computer graphics.
\end{IEEEbiographynophoto}

\begin{IEEEbiographynophoto}{Stephanie Tan} is a PhD candidate at the Pattern Recognition and Bioinformatics Group at Delft University of Technology. She received her BS (2015) from California Institute of Technology and Msc (2017) from Imperial College London. Her research interests are multimodal machine learning with an emphasis on human pose estimation methods, and social scene analysis using social signal processing and computer vision.
\end{IEEEbiographynophoto}

\begin{IEEEbiographynophoto}{Ekin Gedik} completed his PhD degree in the Pattern Recognition and Bioinformatics Group of Delft University of Technology. His research interests include but are not limited to social behaviour analysis, wearable sensing, affective computing and pattern recognition. He focused on analysis and detection of social behaviours, interaction and their connection to various social phenomena.
\end{IEEEbiographynophoto}

\begin{IEEEbiographynophoto}{Laura Cabrera-Quiros} is an Assistant Professor at the Costa Rican Institute of Technology (Instituto Tecnológico de Costa Rica), working in the Electronics Engineering department. Her research focuses on the use of machine learning and non-invasive technologies (e.g. wearable and embedded devices, cameras, physiological sensors) to understand human behavior, monitor health, and improve people’s quality of life.
\end{IEEEbiographynophoto}

\begin{IEEEbiographynophoto}{Hayley Hung} is an Associate Professor in the Socially Perceptive Computing Lab at TU Delft, The Netherlands, where she works since 2013. Between 2010-2013 she held a Marie Curie Fellowship at the Intelligent Systems Lab at the University of Amsterdam. Between 2007-2010 she was a post-doctoral researcher at IDIAP Research Institute in Switzerland. She obtained her PhD in Computer Vision from Queen Mary University of London in 2007. Her research interests are social computing, social
signal processing, computer vision, and machine learning.
\end{IEEEbiographynophoto}




\clearpage
\appendices

\section{Network details}\label{app:seg_method_details}

\begin{figure}[ht!]
\centering
\begin{subfigure}{0.49\textwidth}
  \centering
  \includegraphics[width=\columnwidth]{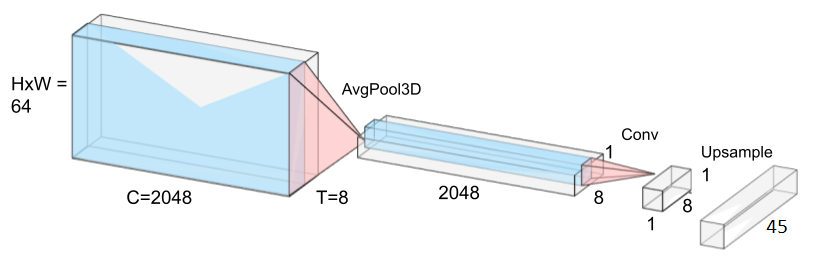}
  \caption{Time series ResNet head (acceleration modality).}
  \label{fig:seghead:accel}
\end{subfigure}

\begin{subfigure}{.49\textwidth}
  \centering
  \includegraphics[width=\columnwidth]{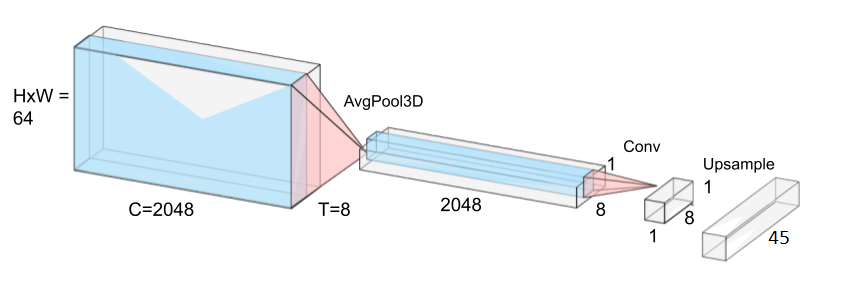}
  \caption{Video ResNet (slow model) head.}
  \label{fig:seghead:video}
\end{subfigure}
\caption{Segmentation heads for acceleration and video models. The first block represents the feature map before the head of the ResNet model, for each modality method. Subsequent operations pool and convolve over the spatial and channel dimensions, and up-sample the time dimension.}
\label{fig:segheads}
\end{figure}

Figure \ref{fig:segheads} presents the architecture of the segmentation heads used. For all models, we apply pooling and convolution operations over the spatial and channel dimensions, and up-sample the time dimension to the length of the target segmentation mask (60). Output masks are averaged for multimodal methods.

\end{document}